\documentclass[conference]{IEEEtran}
\IEEEoverridecommandlockouts
\usepackage{cite}
\usepackage{algorithmic}
\usepackage{graphicx}
\usepackage{textcomp}
\usepackage{xcolor}

\usepackage[misc]{ifsym}
\usepackage{amsmath}
\usepackage{amssymb}
\usepackage{enumerate}
\usepackage{hyperref}
\usepackage{mathrsfs}
\usepackage{bm}
\usepackage{xspace}
\usepackage{graphicx}
\usepackage{float} 
\usepackage{url}
\usepackage{subcaption}
\usepackage{multirow}
\usepackage{tabularray}

\def\BibTeX{{\rm B\kern-.05em{\sc i\kern-.025em b}\kern-.08em
    T\kern-.1667em\lower.7ex\hbox{E}\kern-.125emX}}
\begin{document}

\title{Extrapolating Prospective Glaucoma Fundus Images through Diffusion in Irregular Longitudinal Sequences}


\author{\IEEEauthorblockN{Zhihao Zhao}
\IEEEauthorblockA{\textit{Technical University of Munich}\\
Munich, Germany \\
zhihao.zhao@tum.de}
\and
\IEEEauthorblockN{Junjie Yang}
\IEEEauthorblockA{\textit{Technical University of Munich}\\
Munich, Germany \\
junjie.yang@tum.de}
\and
\IEEEauthorblockN{Shahrooz Faghihroohi}
\IEEEauthorblockA{\textit{Technical University of Munich}\\
Munich, Germany \\
shahrooz.faghihroohi@tum.de}
\and
\IEEEauthorblockN{Yinzheng Zhao}
\IEEEauthorblockA{\textit{Technical University of Munich}\\
Munich, Germany \\
yinzheng.zhao@tum.de}
\and
\IEEEauthorblockN{Daniel Zapp}
\IEEEauthorblockA{\textit{Technical University of Munich}\\
Munich, Germany \\
daniel.zapp@mri.tum.de}
\and
\IEEEauthorblockN{Kai Huang}
\IEEEauthorblockA{\textit{Sun Yat-Sen University}\\
Guangzhou, China \\
huangk36@mail.sysu.edu.cn}
\and
\IEEEauthorblockN{Nassir Navab}
\IEEEauthorblockA{\textit{Technical University of Munich}\\
Munich, Germany \\
nassir.navab@tum.de}
\and
\IEEEauthorblockN{M.Ali Nasseri}
\IEEEauthorblockA{\textit{Technical University of Munich}\\
Munich, Germany \\
ali.nasseri@mri.tum.de}
}
\maketitle

\begin{abstract}
The utilization of longitudinal datasets for glaucoma progression prediction offers a compelling approach to support early therapeutic interventions. Predominant methodologies in this domain have primarily focused on the direct prediction of glaucoma stage labels from longitudinal datasets. However, such methods may not adequately encapsulate the nuanced developmental trajectory of the disease.
To enhance the diagnostic acumen of medical practitioners, we propose a novel diffusion-based model to predict prospective images by extrapolating from existing longitudinal fundus images of patients.
The methodology delineated in this study distinctively leverages sequences of images as inputs. Subsequently, a time-aligned mask is employed to select a specific year for image generation.
During the training phase, the time-aligned mask resolves the issue of irregular temporal intervals in longitudinal image sequence sampling. Additionally, we utilize a strategy of randomly masking a frame in the sequence to establish the ground truth. This methodology aids the network in continuously acquiring knowledge regarding the internal relationships among the sequences throughout the learning phase. Moreover, the introduction of textual labels is instrumental in categorizing images generated within the sequence.
The empirical findings from the conducted experiments indicate that our proposed model not only effectively generates longitudinal data but also significantly improves the precision of downstream classification tasks.
\end{abstract}


\section{Introduction}
\newcommand{\Name}{{\bm{$DILS$}}\xspace}
The longitudinal data analysis plays a crucial role in forecasting the progression of glaucoma, a critical eye condition \cite{artes2005longitudinal}. The current trend of employing longitudinal data for predicting glaucoma is evolving into a highly popular approach \cite{zhang2023application}. Advances in deep learning have significantly propelled the analysis of longitudinal data, especially with the incorporation of techniques such as Long Short-Term Memory (LSTM) \cite{li2020deepgf} networks and transformers \cite{hu2023glim}. These methods have shown considerable promise in extrapolating the future progression of glaucoma based on existing longitudinal data. While these approaches can predict the future state of glaucoma to a certain extent, their output is typically limited to a status label for the condition. 

Physicians can utilize predicted label results to analyze a patient's glaucoma stages. However, these predictions fall short of providing detailed insights into the specific changes and nuances of glaucoma progression \cite{mary2016retinal}. This limitation highlights the need for a more comprehensive approach that not only forecasts the state of the illness but also clarifies the complex trends in its progression.
The ability to predict the future fundus images of glaucoma patients based on longitudinal data could significantly enhance the diagnosis and analysis of the disease \cite{schuman2020review}. With the evolution of diffusion models, there are now general video generation models that have the capability to learn temporal information within sequences. 
However, current video generation models typically generate videos using an initial image and a prompt \cite{xing2023survey}. Therefore, when utilizing these video generation models, we can only independently utilize each frame from the longitudinal data, limiting our ability to learn the underlying factors influencing disease progression between sequences.
Additionally, it is known that longitudinal image sequences of glaucoma are obtained through irregular time interval sampling. Given the inconsistent sequence lengths and time intervals, predicting outcomes becomes challenging when attempting to input sequences of varying lengths into the model.

In order to fully exploit the value of longitudinal data and address the problems inherent in existing longitudinal data generation models, we introduces a diffusion-based model (\Name) in irregular longitudinal sequences  to extrapolate prospective glaucoma fundus images instead of predicting glaucoma stages, inspired by established video generation models. Our model takes longitudinal sequences and a time-aligned mask as inputs, utilizing the mask to select a specific year. Subsequently, it employs the diffusion model to generate the image frame corresponding to that selected year.
A remarkable challenge, as previously discussed, is the inconsistent sequence lengths and time intervals in longitudinal sequences. To address this, our paper adopts a masking approach to compensate for these temporal discrepancies, thus ensuring input sequence uniformity in the training phase. Additionally, to facilitate a more profound understanding of the sequence's internal dynamics, a random masking strategy is applied to partially obscure the input sequence in training phase. The masked frame is then employed as the ground truth for the generation of image frames.
Furthermore, the study introduces a label conditioning mechanism to enhance flexibility in controlling the categorization of generated images. Through this mechanism, each image frame is infused with data regarding the state of disease, thereby guiding the category of images produced by the diffusion model. This incorporation delineates our innovative approach and its potential to significantly advance the field of predictive modeling in glaucoma management, particularly through the lens of sophisticated image generation technologies.

The major contributions of this work can be summarized as follows. Firstly, we introduce a diffusion-based model to predict the future image of longitudinal sequence data. Subsequently, to tackle the issue of inconsistent sequence lengths and time intervals in longitudinal data, we introduce a time-aligned mask. Finally, for enhanced control over the labels of the generated data, we integrate the data labels as conditions into the diffusion process.

\section{Related work}
\textbf{Video Diffusion Model}
Recent advancements in the realm of image generation have marked a significant stride forward, particularly within the text-to-image (T2I) domain \cite{chen2023controlstyle,zhao2024kldd,sun2023sgdiff}, where existing models are now capable of producing images of remarkably realistic quality. Inspired by the success of T2I models, a surge of work in video generation has emerged \cite{esser2023structure,luo2023videofusion,deng2023mv}. These text-to-video (T2V) models achieve stable video production by synthesizing multiple consistent frames from textual or pictorial cues. Notably, these models predominantly focus on unconditional video generation, aiming to create continuous and visually coherent video sequences from random noise or a predetermined initial state without relying on specific input conditions \cite{croitoru2023diffusion}.

Make-A-Video \cite{singer2022makeavideo} and Imagen Video \cite{ho2022imagen}, respectively use the text-to-image diffusion models of DALL·E 2 and Imagen for large-scale T2V generation. Another method involves constructing T2V models based on pre-trained stable diffusion models, which are then wholly or partially fine-tuned according to video data. 
VideoLDM \cite{blattmann2023align} and AnimateDiff \cite{guo2023animatediff} have extended the application of Stable Diffusion \cite{rombach2022high} to 3D environments, training only the newly added temporal layers. This approach demonstrates their capability to integrate with the weights of personalized T2I models.
SVD \cite{blattmann2023stable} highlights the importance of a meticulously curated pretraining dataset to produce top-tier videos and proposes a methodical curation process for training a robust base model, incorporating captioning and filtering techniques. Subsequently, we investigate the effects of fine-tuning our base model using high-quality data and develop a text-to-video model that rivals closed-source video generation systems.
Additionally, LaVie \cite{wang2023lavie} undertook the task of fine-tuning the entire T2V model on both image and video datasets. SEINE \cite{chen2023seine} introduces a novel approach for generating lengthy videos and facilitating image transitions through the use of stochastic masking.

\textbf{Medical Longitudinal Sequence Analysis}
Longitudinal research plays a key role in understanding disease progression, assessing treatment outcomes, and monitoring patient health status \cite{gerig2016longitudinal}. Conducting such analyses requires advanced computational models with the capability to process and interpret alterations in medical images across various time points. Recent advancements have seen methodologies that generate longitudinal brain images using a single image and corresponding age data, allowing for the creation of images depicting either diseased or healthy states based on labels \cite{oh2022learn}. In \cite{xia2021learning}, Xia et al. proposed a method for synthesizing images of the aging brain without longitudinal data, wherein the model adjusts its output based on the inputted age and health status. This approach employs a sequential embedding mechanism to guide the network in learning the joint distribution of brain images, age, and health status. However, methods based on non-longitudinal data often lack temporal dependency. Moreover, many similar approaches are grounded in Generative Adversarial Networks (GANs).

LDGAN \cite{ning2020ldgan} leverages incomplete longitudinal MRI data to predict multiple clinical scores at future time points. Specifically, it estimates MR images by learning a bidirectional mapping between MRIs at two adjacent time points while simultaneously conducting clinical score prediction. This methodology explicitly encourages task-oriented image synthesis. TR-GAN \cite{fan2024tr} complements missing images in MRI datasets through a transformer structure, while SADM \cite{yoon2023sadm} employs a novel 2D image generation method that inputs a sequence of images into a diffusion model for predicting future images. This innovative design allows for learning temporal dependencies and enables autoregressive generation of image sequences during inference. On the other hand, SIGF \cite{li2020deepgf} and GLIM-Net \cite{hu2023glim} focus on predicting stages of glaucoma in longitudinal glaucoma data. However, their primary focus is on stage prediction rather than forecasting future images.

\section{Methodology}

\begin{figure*}[ht]
	\centering
	\includegraphics[width=.98\textwidth]{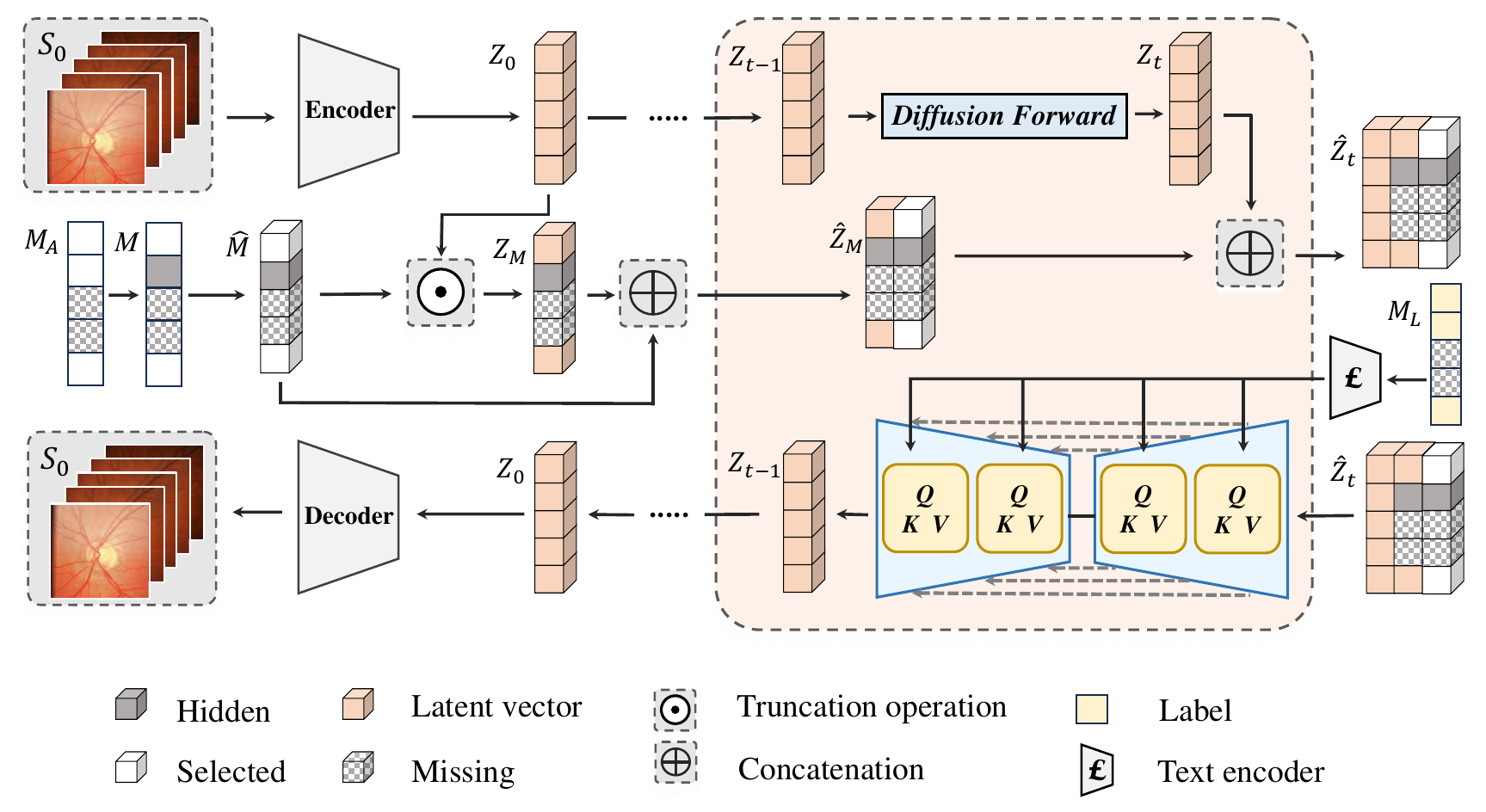}
	\caption{Overview of our proposed \Name, whose architecture is based on a 3D diffusion model. The network's input comprises a sequence of glaucoma images and a time-aligned mask $M_A$. The image sequence is encoded into the latent space $Z_0$. During the training phase, the time-aligned mask $M_A$ randomly selects a frame from the sequence to be hidden,  resulting in a new mask $M$. $\hat{M}$ is obtained by expanding $M$ to the same dimensions as $Z_0$. To better control the attributes of the generated sequence frames, we provide a label mask$M_L$. } 
	\label{flowchart}
\end{figure*}
\subsection{Overview of the Proposed Network}
Figure~\ref{flowchart} illustrates our proposed \Name model, which primarily aims to generate prospective glaucoma fundus images by learning the temporal relationships and disease progression states present within known longitudinal data sequences. The model's input consists of two parts: firstly, our image sequences $S_0$, and secondly, the time-aligned mask $M_A$ corresponding to each frame of the sequence, where a mask value of 1 indicates the selected frames (white cubes) in the sequence, and a value of 255 indicates missing data for the corresponding year (chessboard cubes) in the sequence. The image sequence is initially encoded into the latent space $Z_0 = \mathcal{E}(S_0)$. Then the diffusion forward process adds noise to $Z_0$ to obtain ${Z_1 \ldots Z_T}$. Here, $T$ denotes the number of sample steps used in the diffusion model.
Within the frames where mask $M_A$ has a value of 1, a frame is randomly selected and set to 0 in the new mask $M$, thus  hiding that frame in the training phase to facilitate the model's improved learning of temporal relationships within sequences and the underlying progression of diseases. 
We expand $M$ to the same dimensions as $Z_0$, resulting in mask $\hat{M}$.
To link the original data with the mask, we perform a truncation operation as Equation~\ref{equ:zm} on the original longitudinal sequence $S_0$ and mask $\hat{M}$ to obtain the masked video $Z_M$. 
Furthermore, we concatenate $Z_M$ and mask $M$ together, and then sequentially concatenate this combination $\hat{Z_M}$ with ${Z_1 \ldots Z_T}$ to obtain $\hat{Z_1} \ldots \hat{Z_T}$. The resulting concatenated sequence is then fed into the U-net  of the 3D diffusion model. To enable control over the categories of the generated sequence frames, we introduced a label mask $M_L$. This label mask corresponds to the category labels of each frame, which are binary labels indicating whether the frame is indicative of glaucoma, represented by 0 and 1.

\subsection{Sequence Diffusion Model}
Our sequence generation model is realized through a 3D diffusion model inspired by the video generation model in \cite{chen2023seine}. Given a sequence $S_0 \in \mathbb{R}^{F\times C \times H\times W}$, with each frame representing a year. Here,  $F$ denotes the number of frames in the sequence, $C$ represents the number of channels of the image, and $H\times W$ represent the size of the image. To reduce the computational load on the network, each frame of the input sequence is compressed into a latent space $Z_0 \in \mathbb{R}^{F\times C\times h\times w}$ through a pre-trained VAE encoder \cite{tomczak2018vae}, where in this study $h = H/8, w = W/8$. The entire process of diffusion takes place in these latent spaces. The addition of noise in the diffusion forward process follows a predefined Markov chain $q(Z_t|Z_{t-1})=\mathcal{N}(Z_t;\sqrt{1-\beta_t}Z_{t-1},\beta_tI)$, where $\beta_t$ is a set of predefined decay schedule hyperparameters, $I$ denotes the identity matrix. By iteratively progressing through time steps $t$ from 1 to $T$, we can acquire the complete result of the diffusion forward process, represented as ${Z_1 \ldots Z_T}$. This iterative forward method can be directly obtained from Equation ~\ref{equ:zt}.
\begin{equation}
    \begin{cases}
    Z_t=\sqrt{\alpha_t}Z_0+\sqrt{1-\bar{\alpha}_t}\epsilon,\epsilon\sim\mathcal{N}(0,I)\\[2ex]\bar{\alpha}_t=\prod_{i=1}^t(1-\beta_t)
    \end{cases}
    \label{equ:zt}
\end{equation}

\subsection{Mask-Based Temporal Alignment}
\label{sec:masktempalign}
Due to the  non-uniform sampling times of longitudinal data and the varying lengths of each sequence, maintaining the temporal characteristics of the data and ensuring consistency in the length of input data is challenging. In this study, we adopt a sequence of F-frames $S_0 \in \mathbb{R}^{F\times C \times H\times W}$.
Correspondingly, we utilize a time-aligned mask $M_A \in \mathbb{R}^{F\times 1 \times 1\times 1}$, where a value of 1 in $M_A$ indicates the presence of data for that year, and a value of 255 indicates missing data for that year. This approach allows even non-uniformly sampled data to be represented in our data input format. During the training phase, we randomly obscure a certain frame by binary mask $M \in \mathbb{R}^{F\times 1 \times 1\times 1}$ and then predict the corresponding real frame through the network, because merely predicting the last frame of the image could lead the network to overlook the relationships between frames in the training phrase. After training the network multiple times with random obscuration within the same sequence, it can better capture the temporal characteristics of the sequence and the hidden disease progression states within the sequence frames.

In this paper, we initially broadcast $M$ to $\hat{M} \in \mathbb{R}^{F\times C \times h\times w}$ to achieve the same dimensions as $Z_0 \in \mathbb{R}^{F\times C \times h\times w}$, where $h \times w$ represent the size of latent code in latent space.
We obtain $\hat{M}$ with three values: 0 indicates that the frame is not selected during the training phase, 1 indicates normal selection of the frame, and 255 indicates that the frame does not exist (setting this value to 0 would confuse the network during the image frame generation process, as the 0 value in the mask would correspond both to missing frames and obscured frames). To hide a certain frame during the training phase without affecting the missing frames, this paper defines a truncation operation according to Equation ~\ref{equ:zm}, which implies that we iterate through all elements of the matrix $Z_M$, and select the value of $Z_M$ based on the corresponding value (0 or 1) at matrix $\hat{M}$. Here, $i$ represents the indices of all elements in matrix $Z_M$.
\begin{equation}
    Z_M[i]=
\begin{cases}
    0&\mathrm{if~}\hat{M}[i]=0,\\
    Z_0[i]&\mathrm{if~}\hat{M}[i]=1,\\
    255&\mathrm{otherwise},
    \end{cases}
    (i=[0,1 \ldots ,F\times C\times H\times W])
    \label{equ:zm}
\end{equation}

\subsection{Temporal Attention Module}

\begin{figure}[ht]
\centering
\includegraphics[width=0.3\textwidth]{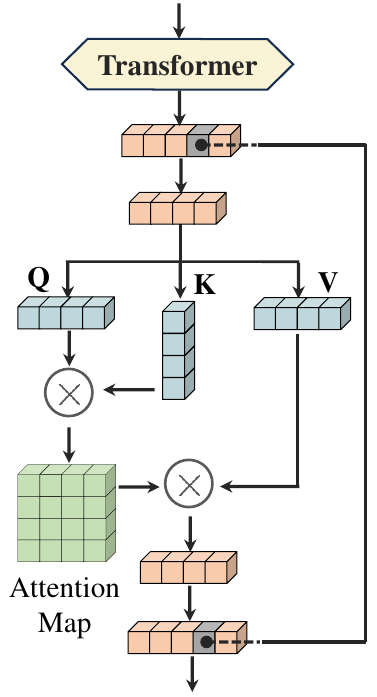}
    \caption{Temporal attention module for capturing information related to disease progression  within the sequence.}
    \label{fig:temporalatt}
\end{figure}

In the diffusion-based UNet model, text vectors and image vectors are interconnected through Cross Attention within the transformer architecture, enabling the control of image generation through text. Following the conventional transformer structure used in the diffusion process, this approach additionally incorporates a temporal attention mechanism to learn the temporal relationships within images. To ensure the model performs well during the inference phase, the hidden state vectors processed by the transformer structure are divided into two parts during the training phase, as illustrated in Figure ~\ref{fig:temporalatt}: one part serves as the known sequence (orange cubes), and the other as the unknown sequence (gray cubes). The positions of the unknown sequences correspond to randomly hidden locations within the time-aligned mask. This setup allows the model to learn the temporal relationships between the known sequences during the training phase, thereby inferring the unknown sequences.

By removing the unknown frames from the sequence according to $\hat{M}$, a new known sequence is formed. This known sequence is then fed into the temporal Attention module, where the attention mechanism captures information related to disease progression inherent within the known sequence. This innovative work highlights the potential of integrating temporal dynamics and attention mechanisms in enhancing the ability of generative models to not only create visually coherent images but also to embed complex temporal sequences and progressions, such as those found in medical imaging studies of disease development. 
Following the thorough capture of temporal information through the attention mechanism, the unknown frames are reintegrated into the output of the Attention module. These frames are then fed into the subsequent transformer stage of the network, thereby maintaining the consistency of the sequence throughout the training phase. This process ensures that the model not only learns the intricate temporal dynamics present within the sequence but also how to seamlessly incorporate previously unknown elements back into the sequence, enriching the model's ability to predict and synthesize future states based on past and present information.

\subsection{Training Methods}
After obtaining $\hat{M}$, $Z_M$, and $Z_t$, we first concatenate $\hat{M}$ and $Z_M$ to get $\hat{Z_M}$. Then, after each forward diffusion to obtain $Z_t$, $Z_t$ is concatenated with $\hat{Z_M}$ to form $\hat{Z_t}$. The reverse process of diffusion involves predicting the noise $\epsilon$ added at each step. We input the concatenated $\hat{Z_t}$ into the U-Net, which outputs the noise added in that iteration. The network model optimizes a minimization loss function as described in Equation~\ref{equ:loss} from paper \cite{nichol2021improved}, thereby yielding the optimal network model $\boldsymbol{\epsilon}_\theta$ in the training phase.

\begin{equation}
\begin{cases}
 Loss=\mathbb{E}\left[||\boldsymbol{\epsilon}-\boldsymbol{\epsilon}_\theta(\hat{Z_t},t)||^2\right]  
\\
\hat{Z}_t=[Z_t;M;Z_M],\tilde{Z}_t\in\mathbb{R}^{n\times(3\times c)\times h\times w}
\end{cases}
\label{equ:loss}
\end{equation}

We also utilized labels to control the generation of sequence frames corresponding to specific labels within the diffusion model. Initially, we convert the labels '0' and '1' of each image frame into the words 'normal' and 'glaucoma', respectively. These are then encoded into corresponding vectors through a text encoder. Subsequently, the vectors for each frame are concatenated to obtain a matrix $M_L$, which is used as a condition that is injected into the U-Net of the diffusion model.

\section{Experiments}

\newsavebox\CBox
\def\textBF#1{\sbox\CBox{#1}\resizebox{\wd\CBox}{\ht\CBox}{\textbf{#1}}}
\subsection{Setup Details}
\subsubsection{Datasets.}
All experiments are carried out on the publicly available  dataset SIGF \cite{li2020deepgf}. 
It comprises 3,671 fundus images, including 405 sequential images from various eyes, averaging nine images per eye, collected between 1986 and 2018, with a minimum of six images for each eye. These images are annotated for positive glaucoma based on criteria such as retinal nerve fiber layer defect, rim loss, and optic disc hemorrhage. Sequences are categorized into time-variant (37 sequences) and time-invariant (368 sequences), where time-variant sequences transition from negative to positive glaucoma diagnoses, while time-invariant ones remain negative. Following the division in \cite{li2020deepgf}, we utilize 300 training, 35 validation, and 70 test images, maintaining similar ratios of sequence types across these sets. The dataset encompasses 264 patients, distributed as 192 in the training set, 23 in the validation set, and 49 in the test set, with a random split at the patient level.

\subsubsection{Evaluation Metric.}
We used the peak signal-to-noise ratio (PSNR) and structural similarity (SSIM) to evaluate the image quality. We also employed Learned Perceptual Image Patch Similarity (LPIPS) \cite{zhang2018perceptual} to measure the distance between the predicted and real data distributions. 
To assess whether the generated images have changed attributes, we employed the attribute matching score (AMS) \cite{chong2021retrieve}, which uses a binary classifier pretrained on glaucoma detection to decide the labels of image pairs. If the labels are the same, we regard them as matching. 
The Vertical Cup-to-Disc Ratio (VCDR) is a crucial metric for assessing the optic disc structure in glaucoma. An increase in VCDR over a period of time indicates a worsening degree of optic nerve damage in glaucoma \cite{czudowska2010incidence}.

\subsubsection{Experimental Settings.}
Our experimental setup utilizes a single NVIDIA RTX A5000 GPU with 24GB of memory and is implemented using PyTorch. The Adam optimizer is employed for optimization, with the initial learning rate for optimizing model parameters established at 1e-5. In the training phase, the images are resized to 256$\times$256, and the dataset is trained over 500 epochs.

\begin{figure*}[ht]
    \centering
    \includegraphics[width=.95\textwidth]{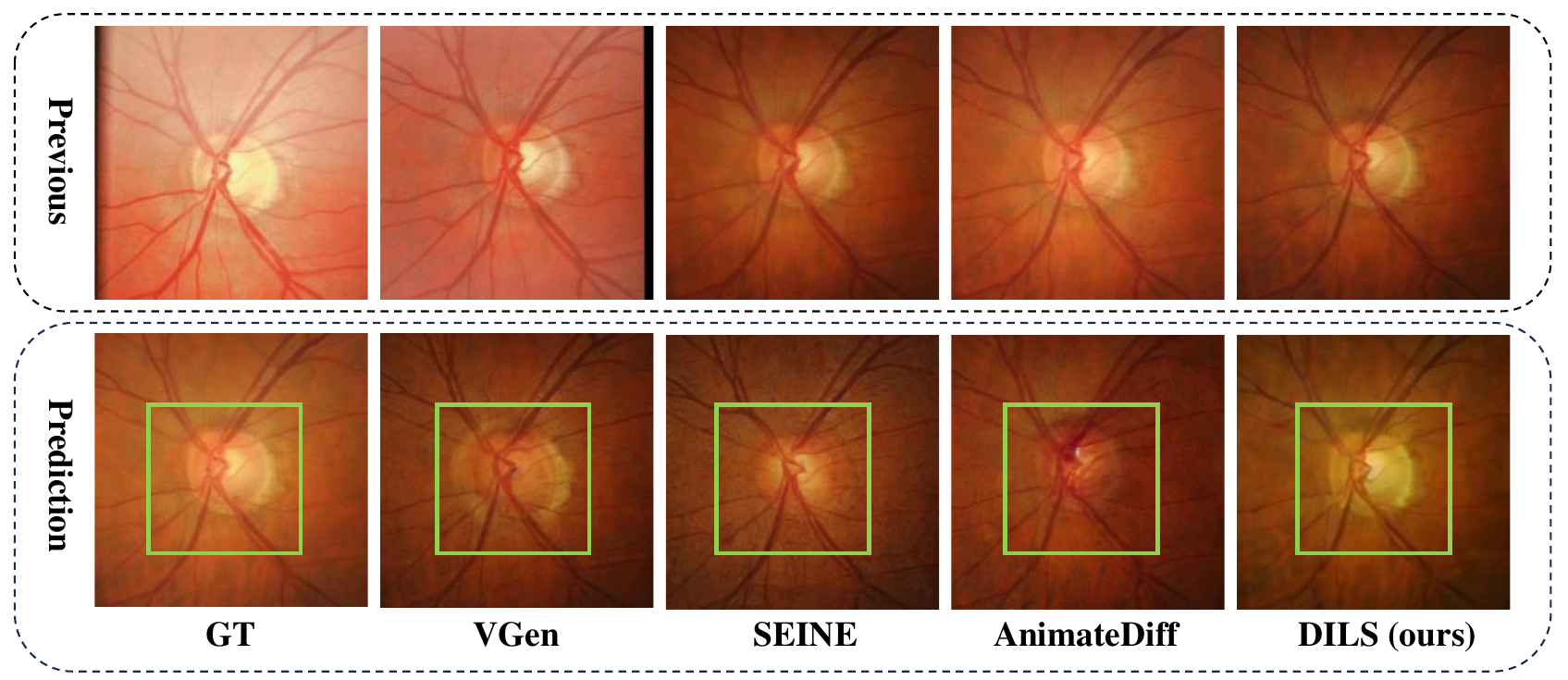}
    \caption{Visualization of glaucoma progression prediction. The first row presents a sequence of images of the patient from various past time periods. The second row shows the ground truth (GT), which represents the current fundus image of the patient; the other images depict the prediction from different methods.}
    \label{fig:prediction}
\end{figure*}

\subsection{Evaluation and Results}

\subsubsection{Results of Prediction}

We first evaluated our method alongside video generation diffusion models in the experiments. For the AnimateDiff \cite{guo2023animatediff} and VGen \cite{zhang2023i2vgen} models, since they can only perform image-to-video operations, we used the last frame of the sequence as the ground truth and the one before the last frame as the input. 
For our \Name model and the SEINE \cite{chen2023seine} model, we also took the last frame as the ground truth and input all other frames directly into the network to predict future frames.
Initially, in Figure~\ref{fig:prediction}, we present the visual outcomes of different methods, noting that the results generated by AnimateDiff and VGen were of acceptable quality. However, as only a single image is used as input without other frames in the sequence, the network can only randomly alter the content of the image based on the prompt. This resulted in noticeable changes in the position and shape of the optic disc in the images. SEINE, with known sequences available, can accurately locate the main feature changes of glaucoma occurring in the optic disc and cup. However, it does not predict the changes in the optic disc according to the trend of glaucoma progression during prediction. Our method can precisely identify the position of the optic disc and, during the generation of predictions, primarily alter the cup-to-disc ratio with only minor changes to the overall optic disc, thus producing a higher correlation between the generated images and the original sequence.

To conduct a human perception evaluation, we generated 100 images each of healthy and glaucoma fundus images, respectively, using different methods. We then invited three  ophthalmologists to assess the quality and labels of the generated images. For the image quality assessment, we randomly paired images generated by other state-of-the-art (SOTA) methods with those generated by our  model  and asked the ophthalmologists to judge which image had higher quality. The evaluations of the doctors are shown in Table~\ref{tab:humanpreference}, where it can be seen that the images generated by our method more closely match the ophthalmologists' expectations.
For the evaluation of the image labels, we mixed the healthy and glaucoma fundus images generated by all methods and asked the doctors to determine whether each image was of a glaucoma fundus. We then computed the AMS metric. The results are presented in Table~\ref{tab:hams}, indicating that our generated images are more consistent with the doctors' perceptions of glaucoma compared to those generated by other SOTA methods.
\begin{table}[h]
\centering
\caption{Human Preference on Generated Image Quality}
\label{tab:humanpreference}
\begin{tblr}{
  vline{2} = {-}{},
  hline{1-3} = {-}{},
}
        & ours$>$AnimateDiff & ours$>$VGen & ours$>$SEINE \\
Quality & 97\%      & 96\% & 93\%     
\end{tblr}
\end{table}

\begin{table}[h]
\centering
\caption{Human Perception Score on Image Attributes}
\label{tab:hams}
\begin{tblr}{
  vline{2} = {-}{},
  hline{1-3} = {-}{},
}
        & AnimateDiff & VGen & SEINE & DILS(ours) \\
AMS & 0.752      & 0.810 & 0.847 & 0.902    
\end{tblr}
\end{table}

To quantitatively assess our results, we employed metrics such as PSNR, SSIM, and LPIPS to quantitatively evaluate the similarity between the generated  frames compared to the previous sequence frames, as shown in Table~\ref{tab:similarity}. Our method can generate an image that is closer to the original sequence, and the AMS score indicates that the label of our generated image is also more consistent with the ground truth.
Additionally, we segmented the optic disc and cup in images from the test set, both glaucomatous and non-glaucomatous, using a fine-tuned Segformer \cite{xie2021segformer} model. We calculated the average vertical cup-to-disc ratio (VCDR) of the generated image frames and the ground truth based on these segmentation results. Furthermore, we classified glaucoma in our generated results using a fine-tuned Swintransformer \cite{liu2021swin} to obtain the AMS score.
According to \cite{czudowska2010incidence}, glaucoma was considered to be present for VCDR values exceeding 0.69 for small discs ($< 2.0 mm^2$), 0.72 for medium-sized discs ($2.0 mm ^2 - 2.7 mm^2$), and 0.76 for large discs ($>2.7 mm^2$).
On the SIGF dataset, we extracted images of glaucomatous and non-glaucomatous eyes and found the average VCDR to be 0.756 for glaucomatous eyes and 0.531 for normal eyes. 
The VCDR results displayed in the table show that the average VCDR of the images generated by our model is closer to the true distribution of the SIGF dataset. Although the VCDR of AnimateDiff on the Normal dataset is closer to the standard for non-glaucoma, it deviates from the distribution of the SIGF dataset. Furthermore, based on its performance on the Glaucoma dataset, it can be observed that its overall result is decreased VCDR, indicating that AnimateDiff does not capture the potential changes of glaucoma.
The results in the table indicate that our method outperforms others in similarity metrics, demonstrating our method's superior ability to identify the inherent relationship between the generated images and the existing sequence.

\begin{table}[h]
\centering
\caption{Comparison of the image generation quality among SOTA video diffusion methods.}
\label{tab:similarity}
\resizebox{0.45\textwidth}{!}{%
\begin{tblr}{
  cell{1}{1} = {r=3}{},
  cell{1}{2} = {c=4}{c},
  cell{1}{6} = {c=2}{c},
  cell{2}{2} = {c=4}{},
  vline{2,6} = {1}{},
  vline{2,6,7} = {2-7}{},
  hline{1,4,7,8} = {-}{},
  hline{2-3} = {2-7}{},
}
Methods       & Similarity &      &    &     & Attribute & \\
              & All        &      &    &     & Glaucoma  & Normal  \\
              & PSNR $\uparrow$       & SSIM $\uparrow$ & LPIPS $\downarrow$ & AMS $\uparrow$  & VCDR   & VCDR \\
AnimateDiff \cite{guo2023animatediff}   & 14.845       & 0.638 & 2.456    & 0.886      & 0.530    & 0.446 \\
VGen  \cite{zhang2023i2vgen}        & 16.388       & 0.690 & 2.203    & 0.928      & 0.703    & 0.685 \\
SEINE \cite{chen2023seine}        & 16.639       & 0.682 & 2.221    & 0.928      & 0.606    & 0.637 \\
DILS (ours)          & 17.379       & 0.703 & 2.118    & 0.957      & 0.743    & 0.560 
\end{tblr}
}
\end{table}

\subsubsection{Downstream Classification}

To further extend the application of our method and evaluate the generated results, we augmented the glaucoma dataset using outcomes produced by our model. In the SIGF dataset, the examples of glaucoma and non-glaucoma are extremely imbalanced, with 37 cases of glaucoma patients and 368 cases of non-glaucoma patients. Some glaucoma classification algorithms are based on single-image classification, while others can predict future glaucoma status based on existing patient image sequences. In this paper, for single-image classification methods, we employed VGG and SwinTransformer, and for sequence-based glaucoma prediction, we used DeepGF \cite{li2020deepgf} and GLIM \cite{hu2023glim} models. To augment the experimental data, we used retinal image sequences from healthy individuals, setting the first five frames as input and controlling the network to generate glaucoma images by setting the label mask for the sixth frame position to 1. We augmented data from 150 healthy retinal sequences. After augmentation, there were 187 sequences for glaucoma and 218 sequences for non-glaucoma. The test dataset remained unchanged, as described in the Datasets section, with the augmented sequences only used in the training set. The accuracy (Acc) improvements of different algorithms after data augmentation were shown in Table~\ref{tab:augmentation}. The experimental results indicate that using various methods for data augmentation enhances the accuracy of glaucoma detection. Our method, compared to not employing any data augmentation, improved the accuracy by over 5\%. Specifically, for the sequence prediction model, our augmentation  increased the accuracy by nearly 7\%. The results also show that the impact of network models on accuracy prediction is more evident than data augmentation. Currently, the transformer-based GLIM-Net outperforms the LSTM-based DeepGF method, and even after data augmentation, the DeepGF method still lags behind GLIM-Net.

\begin{table}[h]
\centering
\caption{The accuracy (Acc) of downstream classification task through data augmentation.}
\label{tab:augmentation}
\resizebox{0.45\textwidth}{!}{%
\begin{tblr}{
  cell{1}{1} = {r=2}{},
  cell{1}{2} = {c=2}{c},
  cell{1}{4} = {c=2}{c},
  vline{2,4} = {-}{},
  hline{2} = {2-6}{},
  hline{1,3,7,8} = {-}{},
}
Methods       & Image based  &                  & Sequence  &      \\
              & VGG-16       & SwinTransformer  & DeepGF    & GLIM-Net \\
None          & 0.697        & 0.847            & 0.771     & 0.871 \\
AnimateDiff\cite{guo2023animatediff}   & 0.703        & 0.852            & 0.790     & 0.885 \\
VGen\cite{zhang2023i2vgen}          & 0.707        & 0.854            & 0.787     & 0.886 \\
SEINE\cite{chen2023seine}         & 0.734        & 0.872            & 0.802     & 0.897 \\
DILS          & 0.760        & 0.895            & 0.842     & 0.928 
\end{tblr}
}
\end{table}

\subsubsection{Model parameters Analysis}
To gain a direct understanding of the model's details, we quantified the number of parameters and Multiply–Accumulate Operations (MACs) across different modules of the model in Table~\ref{tab:params}. The number of parameters serves as a measure of the model's size, while MACs provide an assessment of the model's computational efficiency. The Text\_encoder, VAE\_encoder, and VAE\_decoder in the table are pre-trained models and do not require parameter updates. Our DILS model uses 903.13 million parameters, which is only 5.07\% more than the original 2D Diffusion model's 859.52 million parameters, enhancing the capabilities for longitudinal prediction and irregular data processing. 
\begin{table}[h]
\centering
\caption{Model Params and Macs}
\label{tab:params}
\resizebox{0.45\textwidth}{!}{%
\begin{tblr}{
  vline{2-3} = {-}{},
  hline{1-2,6} = {-}{},
}
Modules       & Params(M) & MACs(M) \\
Text\_encoder & 123.13    & 6782    \\
VAE\_encoder  & 34.16     & 566371  \\
DILS\_Diffusion          & 909.13    & 940174  \\
VAE\_decoder  & 49.49     & 1271959 
\end{tblr}
}
\end{table}
\subsubsection{Ablation Study.}
\begin{figure}[h]

    \includegraphics[width=0.46\textwidth]{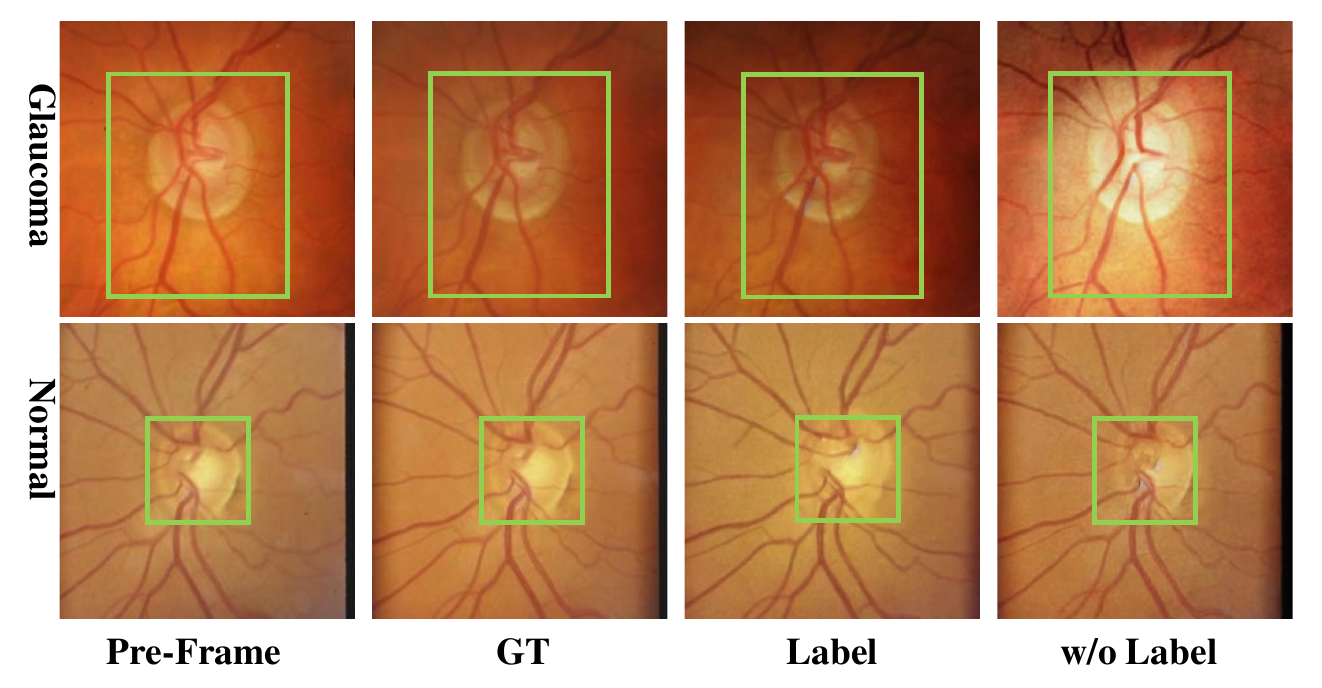}
    \caption{Results of prediction with and without label mask.}
    \label{fig:labelmask}
\end{figure}

To verify whether the label mask plays a role in the generation process of sequence frames and whether the network model can still learn the temporal relationship between sequences and the potential progression of glaucoma without the label mask, we employed a method where the label mask was set to zero, indicating that the label was not used as a condition for image generation. The results generated by different methods were shown in Figure~\ref{fig:labelmask}, with the image quality generated by different methods being roughly equivalent. The quantitative results were presented in Table~\ref{tab:labelmask}, where we only quantified the impact of the label mask on the generated attributes.
From Figure~\ref{fig:labelmask} and Table~\ref{tab:labelmask}, it can be observed that when label masks are not applied, the generated results exhibit greater randomness. However, when label masks are utilized, the generated results become more consistent with the preceding frames of the same label.

\begin{table}[h]
  \centering
  \caption{Measurement of  label mask}
  \label{tab:labelmask}
  \resizebox{.45\textwidth}{!}{%
  \begin{tblr}{
    cell{1}{1} = {r=2}{},
    cell{1}{2} = {c=3}{},
    cell{1}{5} = {c=3}{},
    vline{2,5} = {1-5}{},
    hline{1,3,5} = {-}{},
    hline{2} = {2-8}{},
  }
                & Glaucoma  &      &       & Normal & &      \\
                &LPIPS($\downarrow$)   & AMS($\uparrow$)     & VCDR  & LPIPS($\downarrow$)  & AMS($\uparrow$)     & VCDR \\
  w/o Label     &2.131       & 0.714   & 0.701 & 2.105      &0.921   & 0.603 \\
  w Label       &2.127       & 0.857   & 0.743 & 2.109      &0.968   & 0.560 
  \end{tblr}
  }

\end{table}

\section{Conclusions}
In this paper, we introduce a method to analyze longitudinal data using a diffusion model to monitor the progression of glaucoma. By focusing on the generation of prospective fundus images, our methodology offers a forward-looking perspective on the disease's development, providing insights that were previously unattainable. Through the innovative application of sequence alignment masks and label conditioning, we have demonstrated the feasibility of generating high-fidelity representations of future disease progression, thus contributing to a more nuanced understanding and management of glaucoma.

\bibliographystyle{IEEEtran}
\bibliography{main}

\end{document}